\documentclass{article} % For LaTeX2e
\usepackage{iclr2026_conference,times}

\usepackage{amsmath,amsfonts,bm}

\def\eqref#1{equation~\ref{#1}}
\def\1{\bm{1}}

\DeclareMathAlphabet{\mathsfit}{\encodingdefault}{\sfdefault}{m}{sl}
\SetMathAlphabet{\mathsfit}{bold}{\encodingdefault}{\sfdefault}{bx}{n}

\usepackage{hyperref}
\usepackage{url}
\usepackage{graphicx}
\usepackage{booktabs}
\usepackage{multirow} 
\usepackage[table]{xcolor}
\title{UI2V-Bench: An Understanding-based Image-to-video Generation Benchmark}

\author{
  \textbf{Ailing Zhang}\textsuperscript{1,2}\thanks{Equal contribution.},
  \textbf{Lina Lei}\textsuperscript{2,3}\footnotemark[1],
  \textbf{Dehong Kong}\textsuperscript{2,4}\footnotemark[1],
  \textbf{Zhixin Wang}\textsuperscript{2},
  \textbf{Jiaqi Xu}\textsuperscript{2},
  \textbf{Fenglong Song}\textsuperscript{2}\\
  \textbf{Chun-Le Guo}\textsuperscript{3},
  \textbf{Chang Liu}\textsuperscript{5},
  \textbf{Fan Li}\textsuperscript{2}\thanks{Project leader, $^\ddagger$Corresponding author.}~\footnotemark[3],
  \textbf{Jie Chen}\textsuperscript{1}\footnotemark[3]
  \\[2pt]
  \textsuperscript{1}\,Peking University \quad
  \textsuperscript{2}\,Huawei Noah’s Ark Lab \quad
  \textsuperscript{3}\,Nankai University \quad   \\
  \textsuperscript{4}\,Shenzhen Campus of Sun Yat-sen University \quad
  \textsuperscript{5}\,Tsinghua University
}

\iclrfinalcopy % Uncomment for camera-ready version, but NOT for submission.
\begin{document}

\maketitle

\begin{abstract}
Generative diffusion models are developing rapidly and attracting increasing attention due to their wide range of applications. Image-to-Video (I2V) generation has become a major focus in the field of video synthesis. However, existing evaluation benchmarks primarily focus on aspects such as video quality and temporal consistency, while largely overlooking the model's ability to understand the semantics of specific subjects in the input image or to ensure that the generated video aligns with physical laws and human commonsense.
To address this gap, we propose UI2V-Bench, a novel benchmark for evaluating I2V models with a focus on semantic understanding and reasoning. It introduces four primary evaluation dimensions: spatial understanding, attribute binding, category understanding, and reasoning.
To assess these dimensions, we design two evaluation methods based on Multimodal Large Language Models (MLLMs): an instance-level pipeline for fine-grained semantic understanding, and a feedback-based reasoning pipeline that enables step-by-step causal assessment for more accurate evaluation.
UI2V-Bench includes approximately 500 carefully constructed text–image pairs and evaluates a range of both open-source and closed-source I2V models across all defined dimensions. We further incorporate human evaluations, which show strong alignment with the proposed MLLM-based metrics.
Overall, UI2V-Bench fills a critical gap in I2V evaluation by emphasizing semantic comprehension and reasoning ability, offering a robust framework and dataset to support future research and model development in the field.
\end{abstract}

\section{Introduction}
\label{sec:intro}
 Diffusion models~\cite{ho2020denoising, song2020denoising, dhariwal2021diffusion} have recently achieved remarkable success in image generation and have been rapidly extended to video generation. Early video diffusion models~\cite{wan2025wan, kong2024hunyuanvideo, yang2024cogvideox} primarily focus on the text-to-video (T2V) task. However, in practical applications, the image-to-video (I2V) task is more prevalent and useful, as it takes both a textual prompt and an image as inputs. This not only enhances controllability but also makes the generation process more intuitive for users. With growing demand, I2V models have proliferated rapidly, becoming a major focus of current video generation research.

Despite this progress, evaluation metrics for the I2V task remain limited. Existing benchmarks, such as AIGCBench~\cite{fan2024aigcbench} and AnimateBench~\cite{zhang2024pia}, primarily focus on video quality, temporal consistency, and motion smoothness. However, since I2V models take an image as an additional input, they must not only produce visually high-quality videos but also accurately interpret the semantic content of the input image—an evaluation aspect often overlooked in current benchmarks.
Beyond semantic comprehension, a model’s reasoning ability is also crucial for generating logically coherent videos. It is important to assess whether the model can infer and synthesize events that conform to physical laws and common human understanding—for example, recognizing that \textit{“pulling a trigger”} implies \textit{“a bullet being fired”}. Overall, current evaluation protocols rarely assess such fine-grained semantics or implicit world knowledge, which limits their effectiveness in guiding meaningful model improvements.

\begin{figure}[tbp] 
    \centering
    \includegraphics[width=0.95\linewidth]{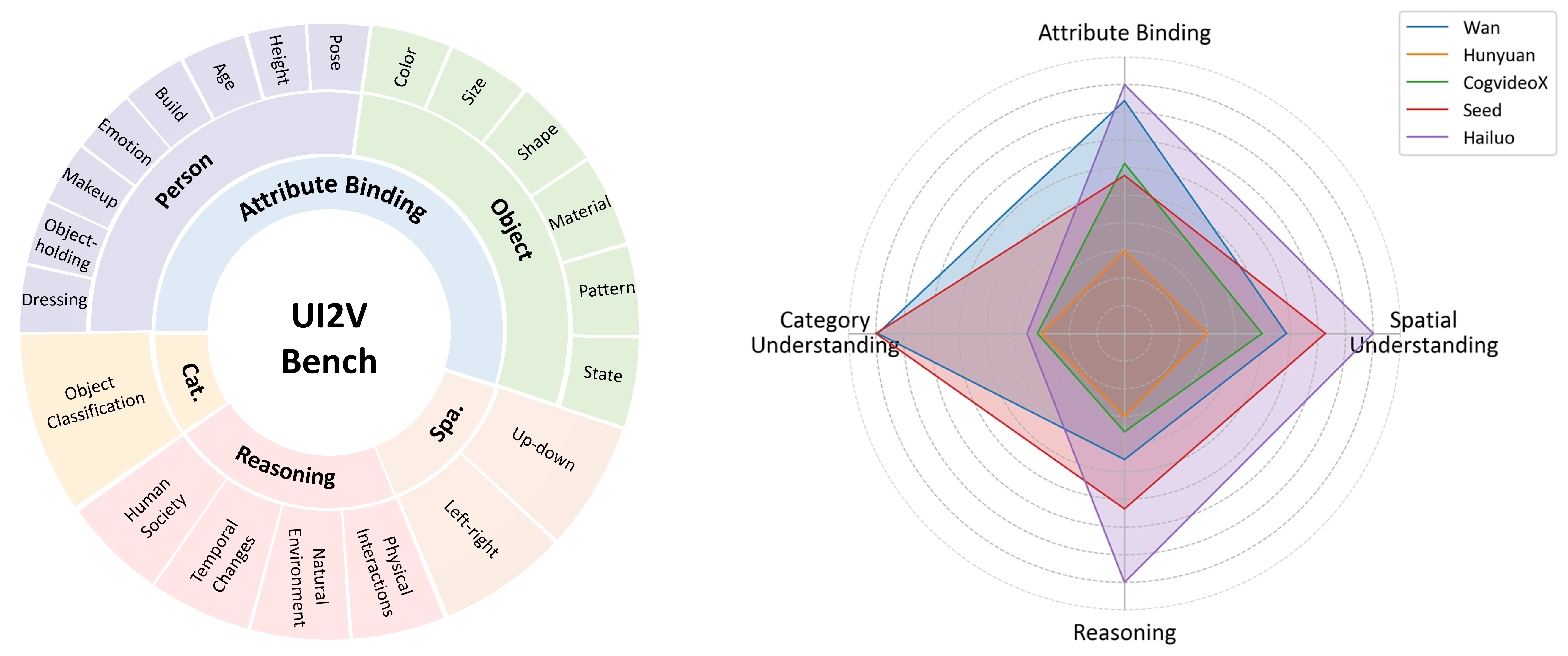} % 调整宽度为行宽的80%
    \caption{Overview of UI2V-Bench. Our understanding-based I2V benchmark consists of $4$ dimensions: spatial understanding, attribute binding, category understanding and reasoning. We evaluate these four capabilities across five representative I2V models. Cat. represents category understanding dimension. Spa. represents spatial understanding.} % 图片标题
    \label{fig:ring} % 图片标签，用于交叉引用
\end{figure}

To address these limitations, we propose UI2V-Bench, an understanding-based benchmark for evaluating image-to-video (I2V) generation models. This framework comprehensively assesses a model’s ability to understand the semantic content of input images and to perform logical reasoning. It consists of four primary evaluation dimensions: spatial understanding, attribute binding, category understanding, and reasoning. As shown in Figure~\ref{fig:ring}, these dimensions are further divided into $9$ aspects and $19$ fine-grained sub-dimensions. Spatial understanding examines how well the model perceives spatial relationships in the input image, such as left–right or up–down positioning. Attribute binding evaluates whether the model can correctly associate distinguishing attributes with specific subjects, including both people and objects. Category understanding measures the ability to differentiate among multiple object categories. Reasoning focuses on the model's capacity for causal inference across various domains, including human society, physical interactions, temporal changes, and the natural environment.

To evaluate these capabilities, we design two MLLM-based evaluation methods. For the three semantic understanding dimensions—spatial understanding, attribute binding, and category understanding—we develop an instance-level evaluation pipeline that supports fine-grained perceptual judgment, as illustrated in Figure~\ref{fig:pipeline1}. For the reasoning dimension, which is often overlooked in existing benchmarks, we propose a feedback-based pipeline that guides MLLMs through a chain of reasoning steps to ensure more accurate assessments, as shown in Figure~\ref{fig:reason2}.

In addition, the UI2V benchmark consists of approximately $500$ carefully designed text–image pairs and evaluates a range of open-source models (e.g., Wan2.1~\cite{wan2025wan}, CogVideoX~\cite{yang2024cogvideox}, Hunyuan-video~\cite{kong2024hunyuanvideo}) and closed-source models (e.g., SeedDance\footnote{SeedDance 1.0 pro, https://www.volcengine.com/}~\cite{gao2025seedance},Hailuo\footnote{Hailuo2.0, https://hailuoai.com/}) across all defined dimensions. Human evaluators also assessed the generated videos, and the results show strong alignment between our evaluation scores and human judgments.

In summary, UI2V-Bench fills a critical gap in the evaluation of I2V models by emphasizing semantic and reasoning capabilities. It provides a systematic and reliable reference standard for guiding the optimization and improvement of future I2V models. We will release the dataset, evaluation suite to facilitate further research and development of I2V models in the community.

\section{Related Work}
\label{sec:related work}
\subsection{Image-to-Video Models}
In recent years, diffusion models(\cite{ho2020denoising};\cite{song2020denoising};\cite{dhariwal2021diffusion};\cite{ho2022video}) have achieved remarkable progress in generative tasks.Early studies focused primarily on image generation(\cite{nichol2021glide};\cite{rombach2022high};\cite{hertz2022prompt};\cite{li2024magiceraser};\cite{kong2025dual}).Building on this, researchers extended diffusion models to video generation, particularly text-to-video (T2V)(\cite{singer2022make};\cite{zhang2023adding};\cite{wu2023tune};\cite{chen2023control};\cite{zheng2024open};\cite{liu2024sora}), which produces temporally coherent video content from natural language descriptions.However, many practical scenarios require not only text but often a single static image.This motivates the study of image-to-video (I2V) generation(\cite{hacohen2024ltx};\cite{yang2024cogvideox};\cite{kong2024hunyuanvideo};\cite{wan2025wan}), which typically incorporates both text and image conditions (TI2V) to preserve the appearance of the input image while introducing plausible temporal dynamics.Despite the rapid development of numerous I2V models, systematic benchmarks for comprehensive performance assessment remain lacking.

\subsection{Evaluation of Video Generation Models}
Evaluation metrics for video generation generally cover visual quality and temporal consistency, with common measures including Inception Score (IS)(\cite{salimans2016improved}) , Learned Perceptual Image Patch Similarity(LPIPS)(\cite{zhang2018unreasonable}), Fréchet Inception Distance (FID)(\cite{heusel2017gans}) , Fréchet Video Distance (FVD)(\cite{unterthiner2018towards}).
In T2V tasks, several benchmarks(\cite{liu2023fetv};\cite{huang2024vbench};\cite{kou2024subjective};\cite{liu2024evalcrafter};\cite{wu2024towards};\cite{sun2025t2v}) have been proposed to systematically evaluate models along multiple dimensions.
% , including video quality, motion coherence, temporal consistency, and text-video alignment
In contrast, benchmarks for I2V remain relatively limited.Existing studies such as AIGCBench(\cite{fan2023aigcbench}) and AnimateBench(\cite{zhang2024pia}) mainly rely on CLIP-based scores, TC-Bench(\cite{feng2024tc}) evaluates Temporal Compositionality, while VBenchI2V(\cite{huang2024vbench++}) assesses consistency by separating the input image and generated video into foreground and background.Nevertheless, these approaches largely overlook the central challenge of I2V: whether the model truly understands the input image and performs reasoning based on it.To bridge this gap, we introduce a new benchmark that provides systematic evaluation along four dimensions: spatial understanding, category understanding, attribute binding, and reasoning.

\section{Bench Construction}
\label{sec:Bench Construction}
In this section, we introduce the main components of UI2VBench. Section~\ref{dimsuite} presents the four primary evaluation dimensions: Spatial Understanding, Attribute Binding, Category Understanding, and Reasoning, where the first three dimensions are together referred to as the semantic understanding dimension of instance-level evaluation. Section~\ref{inputsuite} then details the sources and construction methods of the input images and textual prompts.

\subsection{Dimension Suite}
\label{dimsuite}

\begin{figure}[tbp]
    \centering
    \includegraphics[width=1.0\linewidth]{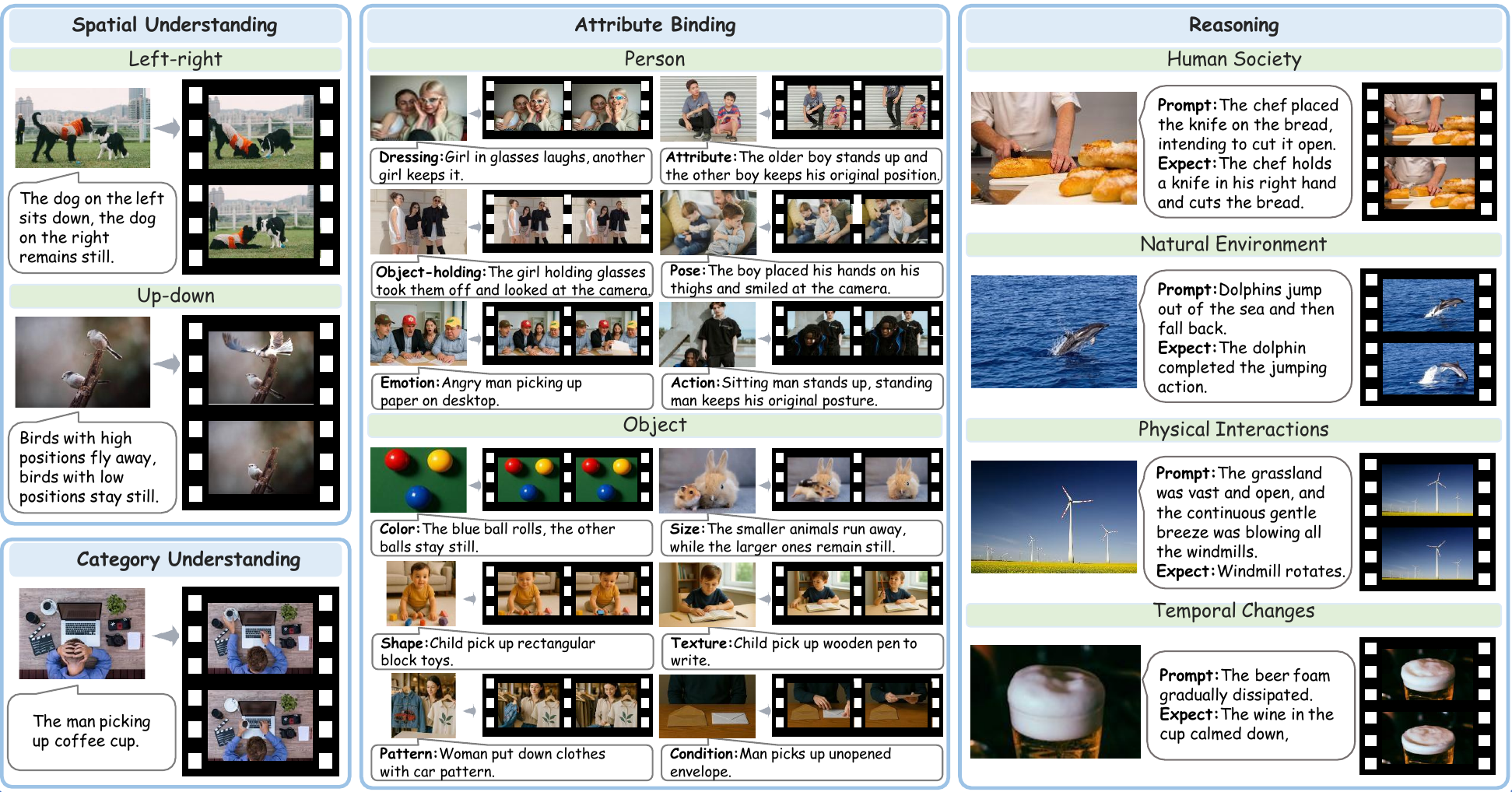} % 调整宽度为行宽的80%
    \caption{Generation cases across evaluation dimensions.For each text–image pair, we show the key video frames of the results corresponding to the $4$ evaluation dimensions: spatial understanding, attribute binding, category understanding, and reasoning.} % 图片标题
    \label{fig:case} % 图片标签，用于交叉引用
\end{figure}

\subsubsection{Spatial Understanding}
Spatial understanding is a critical ability for I2V models. When multiple subjects of the same category appear in the input image with distinct spatial arrangements, users often specify one subject to animate by referring to its spatial location in the textual prompt. This requires the model to first comprehend the spatial relationships in the input image, then accurately localize the target subject, and finally animate it according to the prompt description.

In our benchmark, we construct the spatial understanding dimension using self-designed input cases to evaluate current mainstream I2V models. The spatial relationships considered primarily include vertical and horizontal arrangements. Specifically, when multiple subjects in an image are linearly arranged (e.g., left–right or up–down), the prompt identifies the target subject using spatial ordinal relationships. For example, for an image with two dogs aligned horizontally, a prompt could be: \textit{``The dog on the left sits down, while the dog on the right remains still.''}.  
To prevent spatial information leakage from category-level cues during subject localization, all subjects in this dimension are constrained to the same category. In addition, the number of subjects in our test cases ranges from two to four to enhance diversity and better simulating real-world scenarios.

\subsubsection{Attribute Binding}
In images with multiple subjects, I2V models often need to animate a specific one, guided by textual descriptions of that target (e.g., \textit{``the crying child''}, \textit{``the black table''}). This dimension is designed to evaluate the model’s ability to understand and utilize such attribute-based descriptions for accurate subject localization and animation. Overall, the cases are categorized into two primary types: \textbf{person animation} and \textbf{object animation}, each further divided into sub-dimensions based on attribute types. To ensure that models depend solely on attribute information rather than category differences, all subjects in each case are constrained to the same object category.

\textbf{Person animation.}
Attributes may include age, height, build, dressing, makeup, emotion, pose, and object-holding. 
To specify the target subject based on age, height, or build, we employ comparative or superlative descriptors (e.g., \textit{``the older person''}, \textit{``the taller individual''}, or \textit{``the heaviest''}). 
For dressing, the subject is identified by what they are wearing (e.g., \textit{``the girl in a red dress''}). 
The makeup sub-dimension applies when the target person has distinctive cosmetics in the image (e.g., \textit{``the girl with blue eyeshadow''}). 
For emotion, subjects may be described with expressions (e.g., \textit{ ``a person with a happy expression''}). 
The pose sub-dimension includes descriptions such as \textit{``the boy sitting on the ground''}. 
In the object-holding sub-dimension, the subject is identified by a unique item they are holding, such as \textit{``the man holding a guitar''}.

\textbf{Object animation.}
Specification can be achieved through attributes such as color, size, shape, material, pattern, and state. 
Color is a commonly used attribute for describing objects in images (e.g., \textit{``the red balloon''}). 
Shape may include descriptors such as \textit{triangular}, \textit{cubic}, or \textit{conical}. 
For the material, we construct cases with different materials such as \textit{metal}, \textit{plastic}, \textit{wood}, and \textit{glass}. 
Pattern is used when the object can be uniquely identified by surface markings (e.g., \textit{``the building block with the letter A''}). 
Objects can also be distinguished by their state, for example, differentiating between \textit{an empty cup} and \textit{a cup containing water}.

\subsubsection{Category Understanding}
The cases in the previous two dimensions---
``Spatial Understanding'' and ``Attribute Binding''---are constructed under the constraint that all subjects belong to the same category. 
In contrast, the ``Category Understanding'' dimension is designed to assess whether I2V models can accurately perceive and distinguish multiple object categories present in a single input image. 
Typically, an image contains several common objects from different categories (e.g., \textit{carrot}, \textit{apple}, \textit{banana}), and the model is required to animate the specified object(s) based on the textual prompt.
This dimension evaluates not only the model’s ability to recognize object categories but also its capacity to maintain {identity consistency over time. 
For example, prompts like \textit{``remove the carrot''} or \textit{``the man picking up the coffee cup''} require both precise localization of the target object and temporally consistent animation reflecting its dynamic state.

\subsubsection{Reasoning.}
For the I2V generation task, a model’s reasoning ability refers to its capacity to apply common-sense, logical, and physical knowledge in generating videos that are consistent with real-world dynamics and human interactions. Since existing video benchmarks rarely address this dimension, we introduce four sub-dimensions to comprehensively evaluate it: Human Society, Physical Interactions, Temporal Changes, and Natural Environment.

\textbf{Human Society.}
This dimension evaluates the model's understanding and reasoning capabilities regarding human social behaviors, and relationship. 
% It encompasses inferring the motivations behind human actions, the purposes of tool usage, and the interactive logic within complex social relationships. 
We design representative test cases covering two key aspects: human intentio    n reasoning and social relationship understanding. The former focuses on deducing the purpose of tool usage, behavioral goals, and underlying motivations (e.g. ``\textit{The archer released the bowstring}'') while the latter requires the model to identify and analyze professional associations, familial ties, or social roles between individuals(e.g. ``\textit{The older brother hugs his younger sister}'').

\textbf{Physical Interactions.}
This dimension evaluates the model’s ability to understand basic physical laws, focusing on object interactions and motion changes. A model with physical common sense should predict object dynamics under forces or environmental changes, ensuring that the generated videos follow real-world physical principles. Therefore, we design test cases covering core physical phenomena, including mechanics, fluid dynamics, and gravitational effects, to verify whether the model truly understands fundamental physics and avoids generating content that violates natural laws. (e.g. ``\textit{Let go of the balloon}'')

\textbf{Temporal Changes.}  
This dimension evaluates the model’s ability to understand and predict how objects or scenes evolve over time. Beyond recognizing static attributes such as color, shape, or size, a model with strong reasoning skills should also infer how these attributes change through natural temporal progression. We design representative cases across different time scales, focusing on life processes, environmental changes, and material state transitions. These diverse scenarios comprehensively assess the model’s capacity to reason about dynamic temporal changes (e.g. ``\textit{The banana was left outside for a week}'').

\textbf{Natural Environment.}
This dimension evaluates the model's understanding of natural ecosystems and animal behavioral patterns. ensuring biological and ecological principles. A model with robust natural-world knowledge should accurately reason about the behaviors of animals and plants in specific environments. We collect a large number of images of various species of animals and plants, covering different types, including animal hunting, interactions between animals, plants adapting to environmental changes and so on. (e.g. ``\textit{The bee lands on the flower to collect nectar}'')

\subsection{Input Suite}
\label{inputsuite}
I2V models require an image paired with a tailored text prompt as input to guide the video generation process. 
For the collection of input images, the majority are sourced from open-access websites such as Unsplash\footnote{https://unsplash.com/} and Pexels\footnote{https://pexels.com/}, 
with a small number of synthetic images generated by GPT-4 for some complex test scenes, such as \textit{``Apples arranged in a circle''}.

Specifically, for the \textbf{Spatial Understanding} and \textbf{Attribute Binding} dimensions, where multiple subjects belong to the same category, we ensure that the prompt only drives one object or person, with the other subjects remaining stationary. 
For example, if an image contains three balls, the prompt would specify the blue ball, and only one ball would match this description.
For the \textbf{Category Understanding} dimension, the input image contains various types of common objects, such as vegetables, fruits, and tools. 
For the \textbf{Reasoning} dimension, cases are designed to explore and reason about the hidden world knowledge behind the image. 
In this case, the prompts no longer describe a specific subject to drive its motion. Instead, they provide a precondition for the input image as the ``cause,'' 
guiding the I2V model to generate the corresponding ``effect.''
\begin{figure}[tbp]
    \centering
    \includegraphics[width=0.99\linewidth]{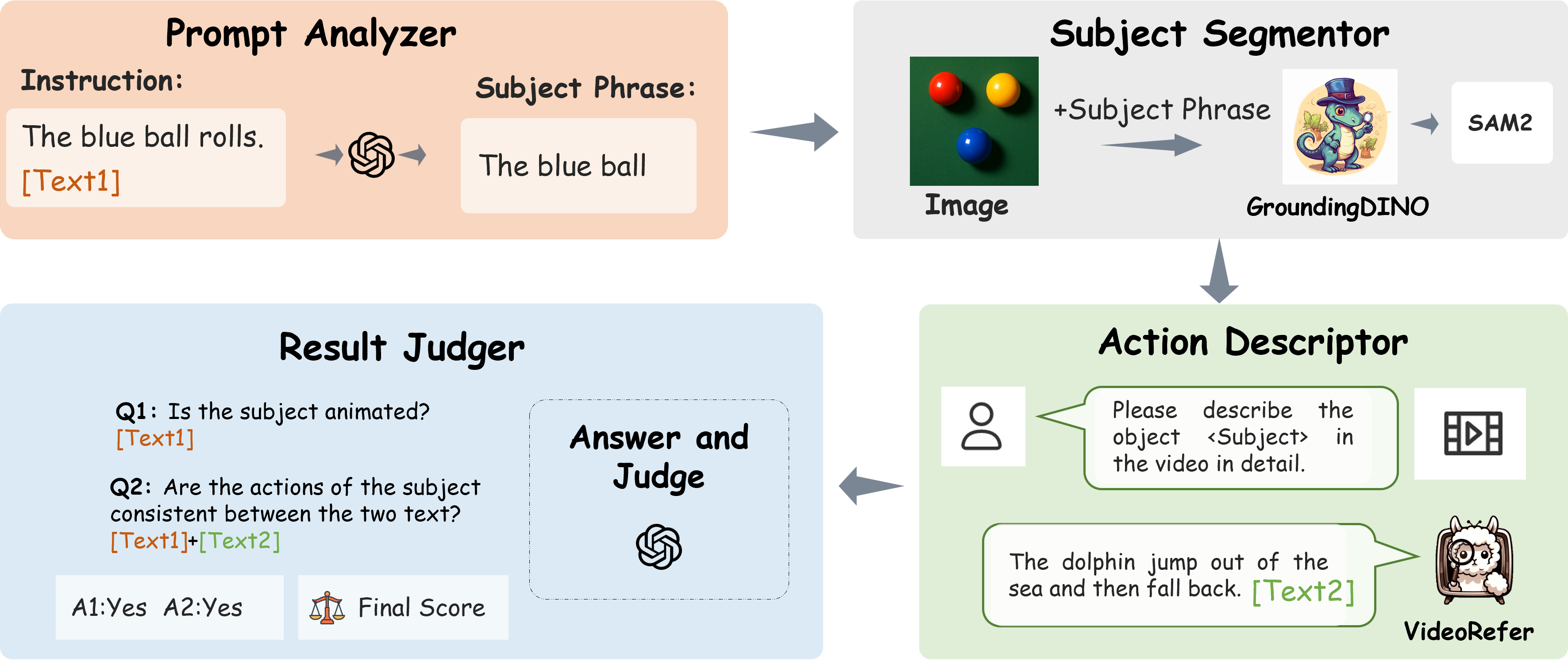} % 调整宽度为行宽的80%
    \caption{Proposed instance-level evaluation framework for semantic understanding.
    Given an input prompt and image, the framework first extracts the subject and action keywords, then obtains the target subject's mask, which is used to generate detailed descriptions. Finally, these descriptions are compared with the original prompt to produce the final evaluation scores.
} % 图片标题 
    \label{fig:pipeline1} % 图片标签，用于交叉引用
\end{figure}

It is challenging to evaluate the image-understanding ability of I2V models, as it requires fine-grained and comprehensive cross-modal understanding. A straightforward yet reliable approach is to employ human evaluators who assess the generated results based on pre-defined rules for each evaluation dimension. However, this method is highly time-consuming and labor-intensive. Therefore, we design two evaluation pipelines based on MLLMs.
\begin{figure}[tbp]
    \centering
    \includegraphics[width=0.99\linewidth]{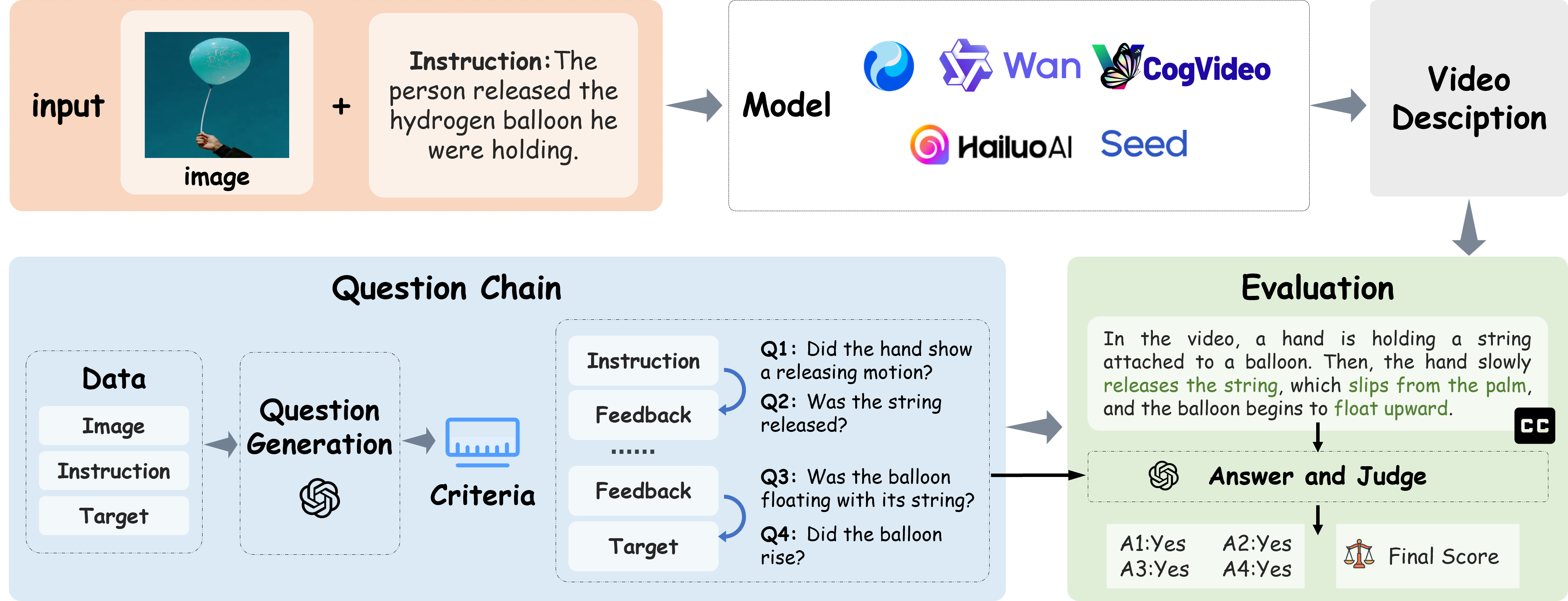} % 调整宽度为行宽的80%
    \caption{Proposed feedback-based evaluation framework for the Reasoning dimension. Given a textual prompt and generated video, the pipeline first produces a detailed video description, then generates a chain of intermediate, observable yes/no questions that guide step-by-step causal validation. Final and intermediate responses are jointly scored to assess the model’s reasoning ability.} % 图片标题
    \label{fig:reason2} % 图片标签，用于交叉引用
\end{figure}
\section{MLLM-based Evaluation Metrics} 
\label{sec:MLLM-based Evaluation Metrics}

\subsection{Instance-level evaluation for sematic understanding}

Current video generation evaluation methods primarily rely on general MLLMs (e.g., LLaVA, Qwen, GPT-4o) to assess output quality through multi-turn QA. 
However, we design an instance-level evaluation pipeline for three semantic understanding dimensions: Spatial Understanding, Attribute Binding, and Category Understanding, 
which similarly require the pipeline to possess fine-grained perceptual abilities. 

As illustrated in Figure~\ref{fig:pipeline1}, the pipeline involves four components: (i) Prompt Analyzer; (ii) Subject Segmentor; (iii) Action Descriptor; (iv) Result Judge.
Specifically, we use MLLMs as the Prompt Analyzer to first extract keywords related to the subject and action, which serve as key information for the subsequent workflow. 
The Subject Segmentor, composed of GroundingDINO~\cite{liu2024grounding} and SAM2~\cite{ravi2024sam2segmentimages}, takes subject keywords as input and extracts the target subject's mask from the input image. The obtained mask is then input into the Action Descriptor, which generates a detailed description of the action and the subject. 
The Action Descriptor utilizes an advanced spatial-temporal object understanding video-LLM VideoRefer ~\cite{yuan2025videorefer},  rather than a common MLLM, which mainly focuses on general scene understanding.
Finally, the Result Judge component utilizes an MLLM to output the final scores by comparing the subject description with the input prompt.

\subsection{Feedback-based Evaluation for Reasoning}

The aforementioned fine-grained, instance-level evaluation methods are particularly suited for the dimensions of semantic understanding. For cases in the Reasoning dimension, greater attention should be given to reasoning about the underlying motivations of the image from the textual prompt input. 
In other words, while common evaluation pipelines focus more on instruction-following capabilities, the Reasoning dimension is primarily concerned with testing the model’s ability for causal inference. 

Evaluating reasoning capabilities faces two challenges. First, MLLMs struggle to accurately identify inconsistencies between text and video due to textual biases~\cite{han2025video}. Second, video content may not align with the target. For example, given an image of a hand holding a pistol and the instruction \textit{``The sniper pulls the trigger''}, the generated video should show the bullet hitting the target, but a bullet is absent in the generated video due to its high speed.

As shown in Figure~\ref{fig:reason2}, we propose a novel feedback-based MLLM evaluation method that generates a chain of progressively validated questions, introducing intermediate feedback to help the MLLM correctly assess the quality of reasoning dimension evaluation cases. The specific steps are as follows:

\begin{enumerate}
    \item Video Description: Generate a detailed description using a video description model (e.g., Tarsier~\cite{yuan2025tarsier2}) to represent the video content for subsequent evaluation.
    \item Question Chain Generation: The LLM generates a set of questions based on preset text prompts and the target. Each question must be a binary judgment (yes/no) or an observable phenomenon that triggers a thought process leading to a feedback result. For example: \textit{``Is there hand motion indicating the trigger is pulled?''}, \textit{``Is there noticeable recoil from the gun?''}, \textit{``Is there a flash of fire?''}, \textit{``Determine if the bullet has been fired''}, etc.
    \item Evaluation Score: Based on the detailed description generated in the first step and the chain of questions generated in the second step, assess the completion of the final result and the intermediate feedback, and output a score.
\end{enumerate}

% \subsection{Overall Metrics for Image-understanding Evaluation}.
% 根据前面

\begin{table*}[t]
\renewcommand{\arraystretch}{1.3}
\centering
\caption{Benchmarking on Reasoning, spatial understanding, category understanding, and attribute binding.} 
\label{ui2v metric}

% --- 第一个表格 ---
\resizebox{\linewidth}{!}{%
\begin{tabular}{lccccc|ccc|c}
\toprule
\multicolumn{1}{c}{\multirow{2}{*}{\textbf{Model}}} & 
\multicolumn{5}{c|}{\textbf{Reasoning}} &
\multicolumn{3}{c|}{\textbf{Spatial Understanding}}   &  \multicolumn{1}{c} {\textbf{Category Understanding}}         \\
              \cmidrule(lr){2-6}\cmidrule(lr){7-9}\cmidrule(lr){10-10}
\multicolumn{1}{c}{}   & HS    &NE  &PI  &TC &Avg.  &Left-right  &Up-down &Avg.  & Object Classification   \\
              \midrule
HunyuanVideo   & 53.85 & 21.47 & 35.21 & 25.83 &34.81  &  23.81 & 24.67 & 24.24 & 20.00  \\
CogvideoX   & 52.18 & 35.90 & 32.50 & 27.67 &37.80  & 30.78   & 36.4 & 33.59 & 20.26 \\
Wan2.1   & 53.51 & 48.98   & 41.46 & 24.92 & 43.18 & 41.72 & 33.80 & 37.76 & 30.04\\
SeedDance  &73.45  & 50.77   & 45.91 & 38.25 &52.77 & 44.04  & 44.77 & 44.41 & 30.15  \\
Hailuo     &71.14  &71.58  &62.76  &70.19  &  67.04  & 49.5 & 55.64 & 52.57 & 20.89 \\ 
\bottomrule
\end{tabular}
}

\resizebox{\linewidth}{!}{%
\begin{tabular}{lccccccccccccc}
% --- 第一行：总标题 "Attribute" ---
\multicolumn{1}{c}{\multirow{3}{*}{\textbf{Model}}} & 
\multicolumn{12}{c}{\textbf{Attribute Binding}} &
\multicolumn{1}{c}{\multirow{3}{*}{}} \\
\cmidrule(lr){2-14}

% --- 第二行：Person 和 Object ---
& \multicolumn{5}{c}{\textbf{Person}} & \multicolumn{6}{c}{\textbf{Object}} & \\
\cmidrule(lr){2-6} \cmidrule(lr){7-14}

% --- 第三行：具体指标 ---
& Dressing & Person-Attribute & Object-holding & Emotion & Person-Action 
& Color & Size & Shape & Textures & Pattern & Condition &Avg. \\
\midrule
HunyuanVideo    & 24.26 & 27.50 & 23.33 & 20.00 & 21.67 & 23.23 & 56.67 & 51.67 & 56.30 & 27.92 & 51.46 &  34.91 \\
CogvideoX    & 27.40 & 42.08 & 41.11 & 22.58 & 20.83 & 29.38 & 52.50 & 46.11 & 69.44 & 35.83 & 53.33 &  40.05  \\
Wan2.1   & 28.97 & 49.58 & 28.33 & 27.67 & 57.50 & 28.33 & 44.17 & 61.48 & 65.74 & 35.45 & 53.96 &  43.74 \\
SeedDance  & 27.39 & 29.17 & 36.23 & 22.37 & 20.00 & 35.40 & 44.52 & 47.69 & 56.96 & 48.33 & 64.55 & 39.33 \\
Hailuo  & 37.37 & 24.52 & 33.67 & 37.86 & 81.67 & 36.44 & 51.41 & 35.33 & 38.67 & 62.78 & 51.88 & 44.69 \\ 
\bottomrule
\end{tabular}
}
\end{table*}

\section{Experiments}
\label{sec:Experiments}

\subsection{Experimental Setup}

\noindent \textbf{Evaluated models}.
For a comprehensive evaluation of the current development in I2V generation, We adopt 3 mainstream open-source I2V models and 2 close-source commercial I2V models for evaluation. Specifically, the open-source models are Wan2.1(\cite{wan2025wan}), HunyuanVideo(\cite{kong2024hunyuanvideo}), CogvideoX(\cite{yang2024cogvideox}). More will be added as they become open-sourced. Besides, we select 2 close-source commercial models: Seedance(\cite{gao2025seedance}), Hailuo, which gain awesome Elo score and high rank on Artificial Analysis Video Arena Leaderboard. The Arena Elo system, adapted from chess, objectively ranks models through anonymous user votes on randomized model matchups.
% todo hailuo参考文献

\noindent \textbf{Implementation details}.
For each sub-ability dimension, videos are generated using the models based on the corresponding prompt suite described in Section \ref{inputsuite}. For the open-source models, we select the most advanced available version of each model for evaluation. For Wan2.1, we employ the 14B version, which yields 5-second videos rendered at 480p and 24 FPS.HunyuanVideo produces 3-second outputs at 720p resolution with a frame rate of 24 FPS. CogVideoX1.5-5B generates 6-second videos at a resolution of 768×1360 and 8 FPS. For the closed-source commercial models, we select versions that balance performance and API cost. Seedance-1.0-pro is employed to synthesize 5-second videos at 580p resolution and 25 FPS, while MinMax-Hailuo-02 is chosen for its superior generation quality, producing 6-second clips at 768p resolution and 24 FPS.

\subsection{Evaluation Metrics}

\begin{table*}[t]
\renewcommand{\arraystretch}{1.3}
\centering
    \caption{Comparisons of video quality and video-condition alignment metrics.} 
\label{quality metric}
\resizebox{\linewidth}{!}{%
\begin{tabular}{lcccccc}
\toprule
\multicolumn{1}{c}{\multirow{2}{*}{\textbf{Model}}} & \multicolumn{3}{c}{\textbf{Video Quality}} & \multicolumn{3}{c} {\textbf{Video-condition alignment}}  \\
              \cmidrule(lr){2-4}\cmidrule(lr){5-7}
\multicolumn{1}{c}{}    &lmage Quality  &Aesthetic Quality   &Motion Smoothness    &Video-text Alignment  &Video-image Similarity  &lmage Understanding (Ours) \\
              \midrule
HunyuanVideo    & 0.7066 & 0.6026  & 0.9941  & 0.2136 & 0.9305 & 28.49  \\
CogvideoX   & 0.7097 & 0.5642 & 0.9887 &  0.2145   & 0.9229 &  32.93 \\
Wan2.1   & 0.7139  & 0.6024  & 0.9865   & 0.2093 &0.9313  &   38.68   \\
SeedDance  & 0.7321 & 0.6080 & 0.9926 & 0.1976  & 0.9009 &  41.67 \\
Hailuo      & 0.7280 & 0.5909 & 0.9943  & 0.2048 & 0.8937 & 46.30 \\

\bottomrule
\end{tabular}
}
\vspace{0pt}
\end{table*}

\noindent \textbf{Our proposed Metrics}.
In our experiments, we evaluate all models using four major metrics: Attribute Binding, Category Understanding, Reason, and Spatial Understanding, as shown in Table \ref{ui2v metric}. HS denotes Human Society, NE denotes Natural Environment, PI denotes Physical Interactions, and TC denotes Temporal Changes. These metrics allow us to quantitatively compare model performance across different reasoning and spatial dimensions. The average of the results from these four dimensions is recorded in Table~\ref{quality metric} as a metric named Image Understanding, reflecting the model's comprehensive capability in image understanding.

\noindent \textbf{Previous Metrics}. 
In addition, we measure several fundamental quality dimensions to ensure the comprehensiveness of our benchmark,as shown in Table \ref{quality metric}. Concretely, we report image quality, aesthetic quality and motion smoothness using VBench(\cite{huang2024vbench}).These metrics complement our evaluation by focusing on essential aspects of perceptual quality and coherence.Furthermore, we include video-text consistency and video-image similarity from AIGCBench(\cite{fan2023aigcbench}), which offers a point of reference for comparing our benchmark with existing I2V evaluations.

\subsection{Quantitative and Qualitative Evaluation}
Table~\ref{ui2v metric} reports the quantitative comparison across four dimensions. (1) Closed-source models generally outperform open-source models.Wan2.1 leads among open-source models, CogVideoX slightly outperforms HunyuanVideo in reasoning, and Hailuo shows a clear advantage over SeedDance, especially on the TC metric.
(2) Their evaluation of generated video quality no longer meets the requirements with fundamental metrics only. For the semantic understanding dimension, it fails to accurately measure how well the generated output implements and responds to fine-grained information in the prompt. For the reasoning dimension, the quality of the generated result is no longer positively correlated with its alignment to the semantic information of the prompt.
The qualitative evaluation shows similar observations.Hailuo exhibits strong reasoning capabilities. For example, given the prompt \textit{``Leave the bananas for a week''}, the model can infer that the bananas will rot after some time,as shown in Appendix Figure~\ref{fig:qua113}.

\begin{figure}[tbp]
    \centering
    \includegraphics[width=0.95\linewidth]{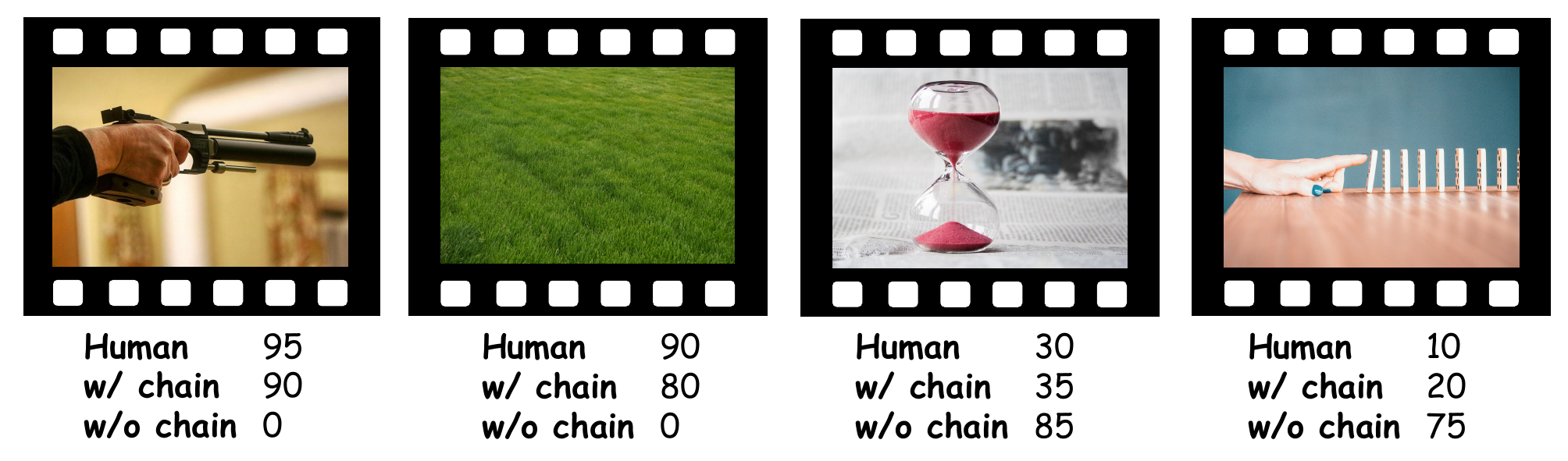} % 调整宽度为行宽的80%
    \caption{Evaluation on reasoning with Question Chain is aligned with human evaluation.} % 图片标题
    
    \label{fig:reason3} % 图片标签，用于交叉引用
\end{figure} 

\subsection{Human Evaluation}
We conduct a human evaluation of the 19 metrics across the four dimensions proposed in this paper. Specifically, we randomly  select samples from our evaluation dataset, where each sample consists of a text prompt, an input image, and generated videos from different models. Evaluators are required to score each sample with simultaneous reference to the input image and text prompt, and each sample is evaluated by at least 8 evaluators. For each generated video, evaluators assign scores for the 19 metrics individually on a 1–5 scale. To reduce inter-evaluator bias, the scores from each user are further normalized before aggregation. Subsequently, we calculate the average score of all evaluators for each metric as the human subjective score.

We calculate Kendall’s tau ($\tau$) and Spearman’s rho ($\rho$) to reveal the similarity between our proposed metric and human evaluation, as shown in Table \ref{human correlation}. Our proposed metrics exhibit  high correlations with human judgments. In particular, the Spatial Understanding and Reasoning metrics achieve the strongest alignment with human evaluation, indicating their ability to capture semantic qualities and implicit reasoning beyond pixel-level measures. These results confirm that our benchmark provides reliable and human-consistent evaluation criteria.

\begin{table*}[t]
\renewcommand{\arraystretch}{1.3}
\centering
\vspace{-1em}
    \caption{Comparisons of human correlation among different dimensions.} 
\label{human correlation}
\resizebox{0.95\linewidth}{!}{%
\begin{tabular}{lcccccccc|cc}
\toprule
\multicolumn{1}{c}{\multirow{2}{*}{\textbf{Metric}}} & \multicolumn{2}{c}{\textbf{Spatial Understanding}} & \multicolumn{2}{c} {\textbf{Attribute Binding}} & \multicolumn{2}{c} {\textbf{Category Understanding}} & \multicolumn{2}{c} {\textbf{Reasoning}} &
\multicolumn{2}{|c} {\textbf{Avg. (ous)}}
\\
              \cmidrule(lr){2-3}\cmidrule(lr){4-5}\cmidrule(lr){6-7}\cmidrule(lr){8-9}\cmidrule(lr){10-11}
   &  $\tau$  &  $\rho$  &  $\tau$  &  $\rho$    &  $\tau$  &  $\rho$  &  $\tau$ & $\rho$ & $\tau$ & $\rho$  \\
              \midrule
Human Correlation   & 0.6052 & 0.7345 & 0.2980 & 0.3566   &0.5729 & 0.6683 & 0.6121 & 0.7329 & 0.5220 & 0.6231 \\

\bottomrule
\end{tabular}
}
\vspace{0pt}
\end{table*}

\section{Conclusions}
\label{sec:Conclusions}
We propose UI2V-Bench, a benchmark for evaluating I2V models in image understanding and prompt responsiveness. It covers semantic understanding (Spatial Understanding, Attribute Binding, and Category Understanding) and implicit reasoning, with automated evaluation pipelines based on MLLMs validated against human perception. We benchmark both open-source and commercial models through quantitative and qualitative analyses, revealing challenges in fine-grained subject–action alignment and in leveraging world knowledge for event prediction. We hope our work will inspire future improvements in the understanding ability of I2V models.

\bibliography{iclr2026_conference}
\bibliographystyle{iclr2026_conference}
\clearpage
\newpage
\appendix

\clearpage
\newpage
\section{MORE DETAILS ON EVALUATION}
\subsection{System Prompt}
\begin{figure}[htbp]
    \centering
    \includegraphics[width=0.88\linewidth]{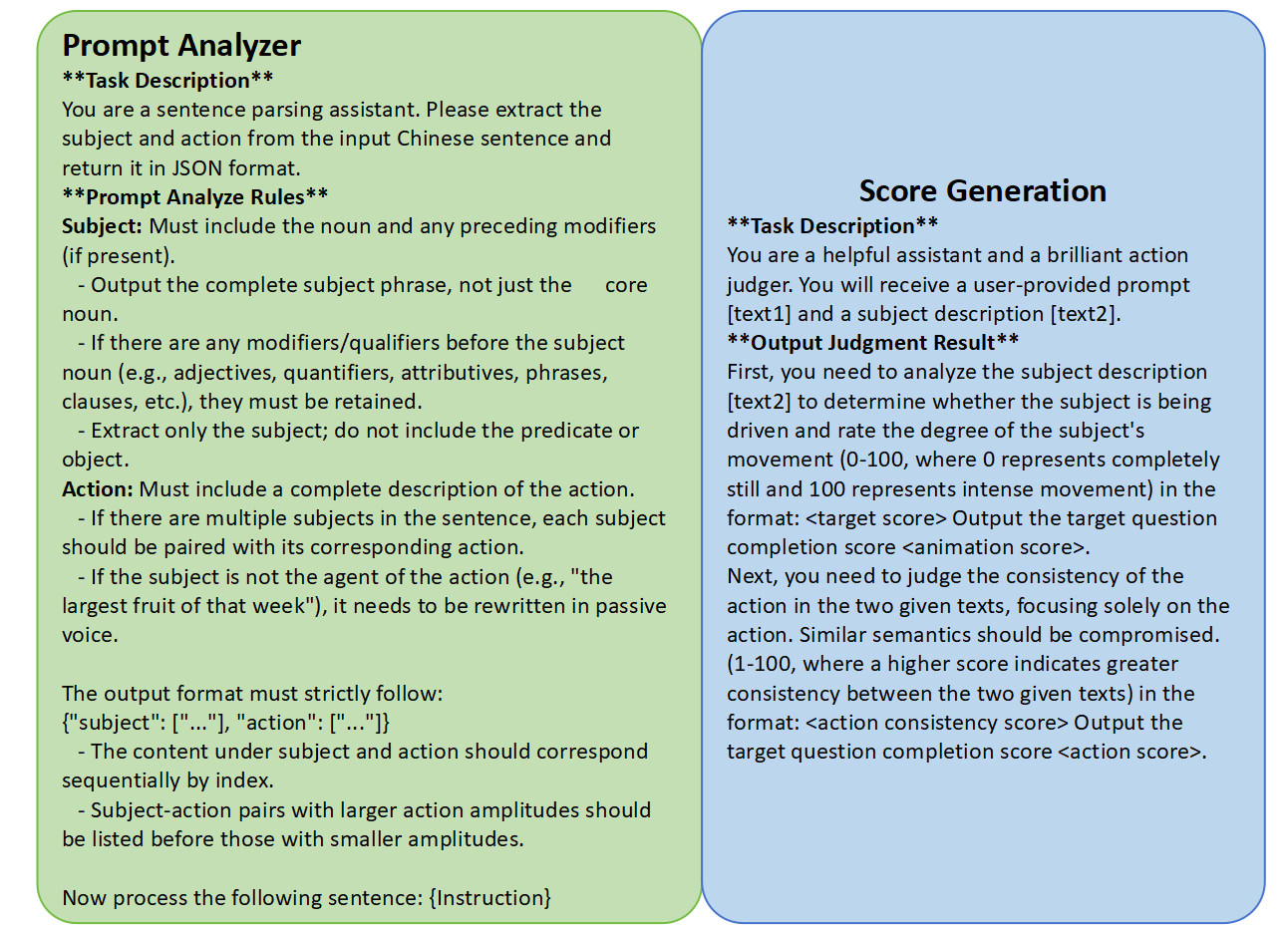} % 调整宽度为行宽的80%
    \caption{system prompt for semantic understanding evaluation} % 图片标题
    \label{fig:reason2} % 图片标签，用于交叉引用
\end{figure} 

% \subsection{EVALUATION FOR REASONING}
% \subsubsection{System Prompt}
\begin{figure}[htbp]
    \centering
    \includegraphics[width=0.88\linewidth]{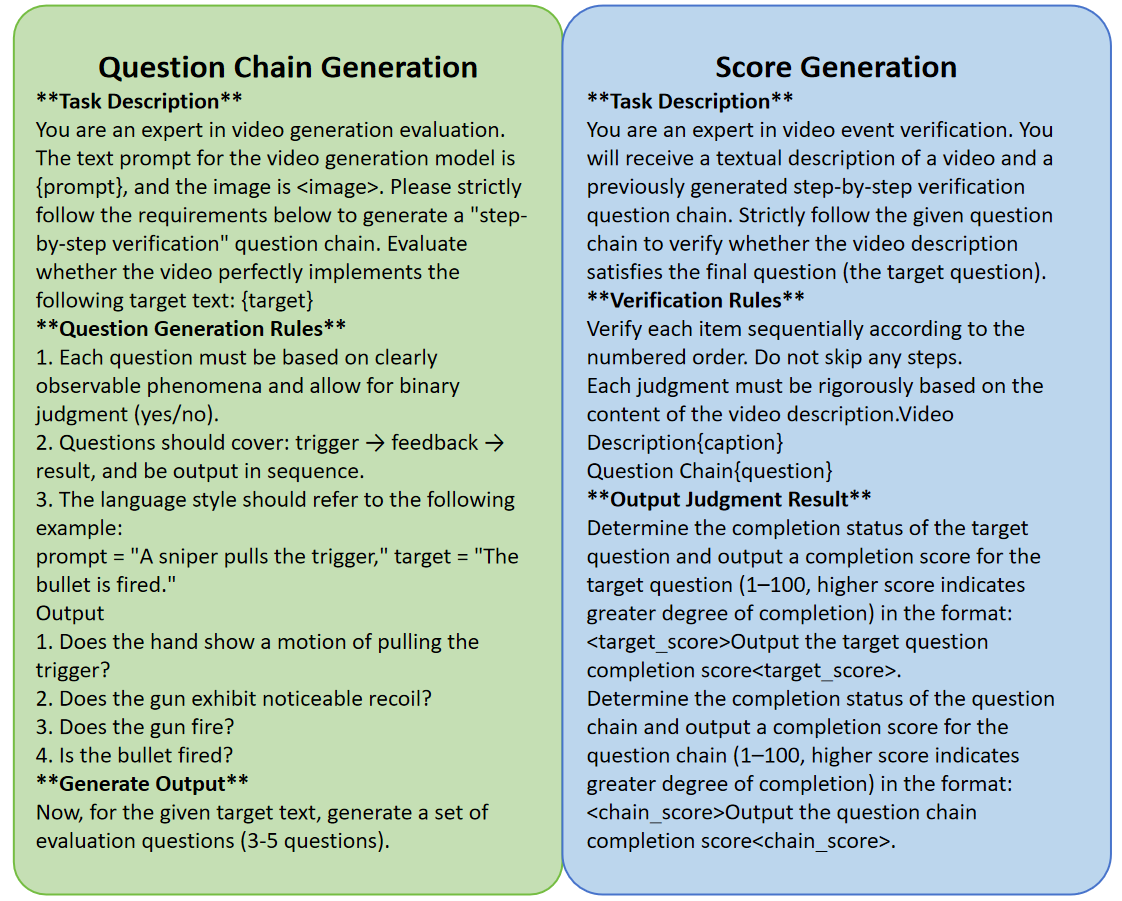} % 调整宽度为行宽的80%
    \caption{system prompt for reasoning evaluation} % 图片标题
    \label{fig:reason2} % 图片标签，用于交叉引用
\end{figure} 

\subsection{Question Chain Details}
This section presents the question chain score (Table \ref{benchmark:metrics1}) and specific question chain examples (Figure \ref{fig:case1},\ref{fig:case2}, \ref{fig:case3}). The score of the problem chain is basically consistent with the final score.
\begin{table*}[ht]
\renewcommand{\arraystretch}{1.3}
\centering
    \caption{Comparisons of Question Chain Score. 
    } 
\label{benchmark:metrics1}
\resizebox{\linewidth}{!}{%
\begin{tabular}{lcccc>{\columncolor{gray!30}}c}
\toprule
\multicolumn{1}{c}{\multirow{2}{*}{\textbf{Model}}} & \multicolumn{5}{c}{\textbf{Reasoning}}          \\
              \cmidrule(lr){2-6}
\multicolumn{1}{c}{}    & Human Society    &Natural Environment  &Physical Interactions  &Temporal Changes &Avg.    \\
              \midrule
Wan2.1   & 57.08 & 52.01   & 45.11 & 27.38  & 46.37   \\

HunyuanVideo   & 56.25 & 31.88 & 40.43 & 29.83 &40.27     \\

CogvideoX    & 52.15 & 43.94 & 38.64 & 32.65 &42.35    \\

SeedDance  &56.62  & 62.45  & 73.03 & 39.00 &58.02   \\

Hailuo   & 79.64   & 68.11 & 65.17 &  73.62  & 69.89     \\ 
\bottomrule
\end{tabular}
}
\vspace{0pt}
\end{table*}

\begin{figure}[htbp]
    \centering
    \includegraphics[width=0.8\linewidth]{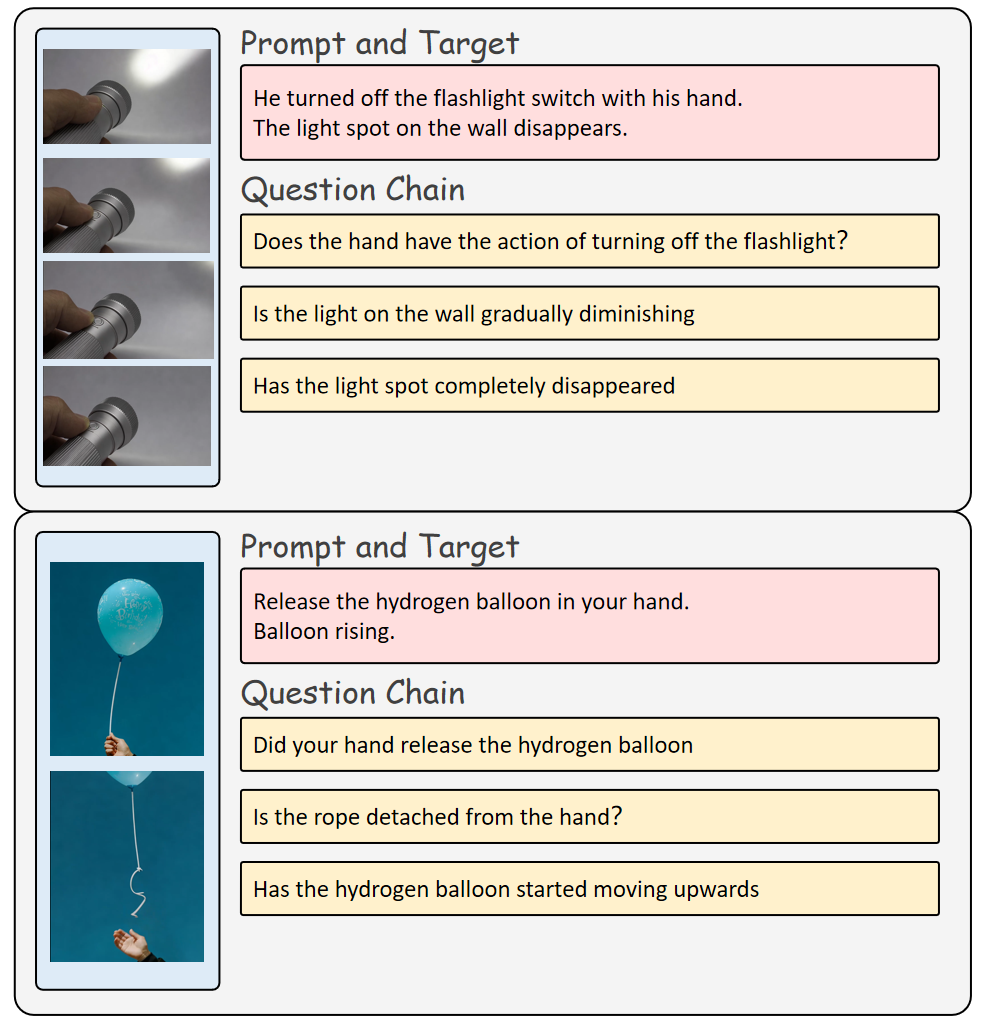} % 调整宽度为行宽的80%
    \caption{Question Chain cases} % 图片标题
    \label{fig:case1} % 图片标签，用于交叉引用
\end{figure}

\clearpage
\newpage

\begin{figure}[htbp]
    \centering
    \includegraphics[width=0.8\linewidth]{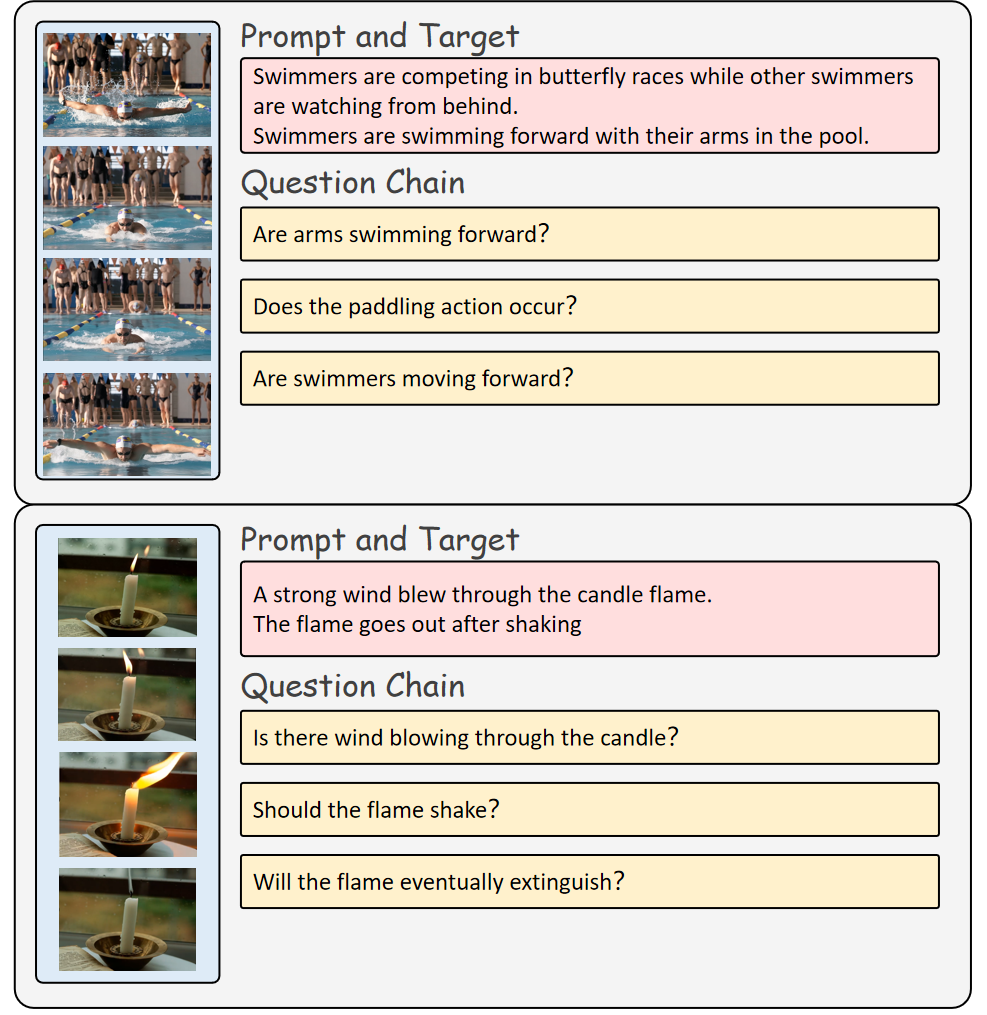} % 调整宽度为行宽的80%
    \caption{Question Chain cases} % 图片标题
    \label{fig:case2} % 图片标签，用于交叉引用
\end{figure}

\clearpage
\newpage

\begin{figure}[htbp]
    \centering
    \includegraphics[width=0.88\linewidth]{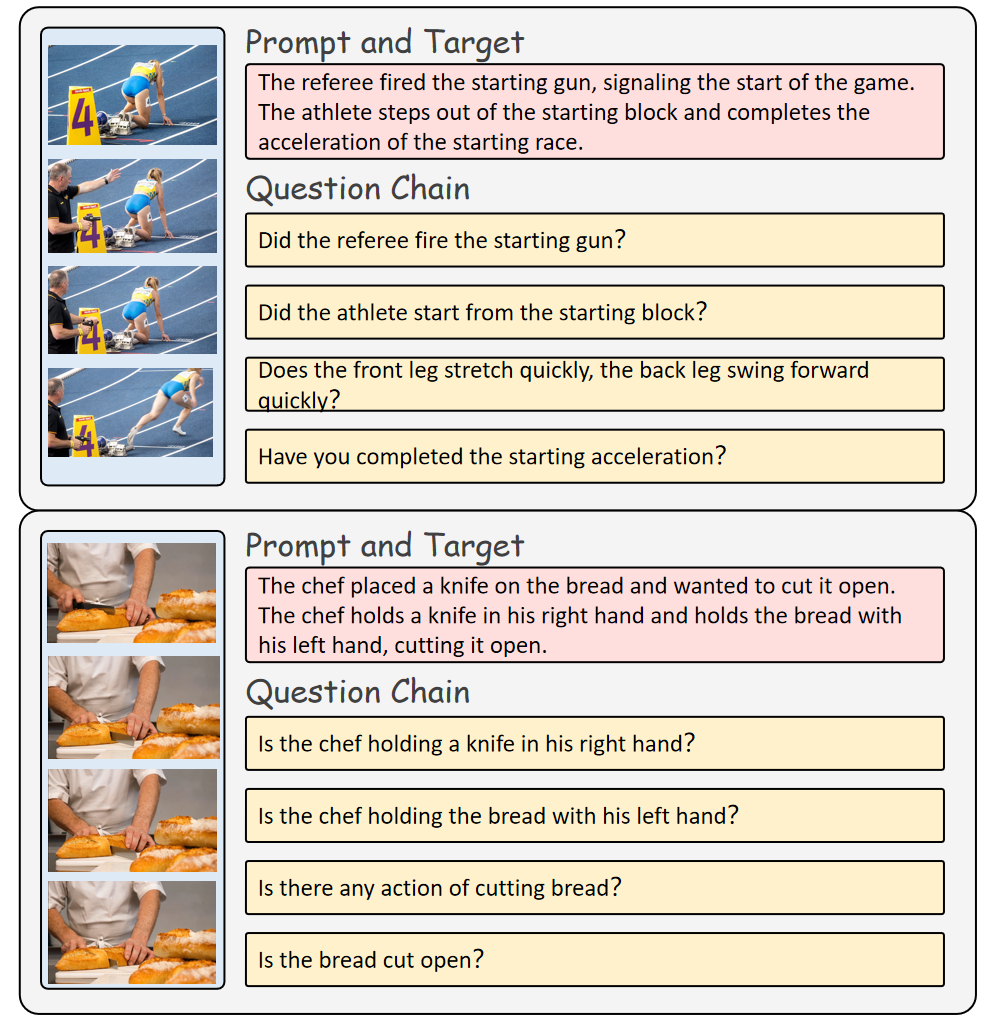} % 调整宽度为行宽的80%
    \caption{Question Chain cases} % 图片标题
    \label{fig:case3} % 图片标签，用于交叉引用
\end{figure}

\section{HUMAN EVALUATION}
% 由于人类评估具有较高的主观性，为了确保不同人的评分一致性，我们提前设定了评分标准。
To assess the correlation of the scores given by UI2VBench with human preferences, we conduct a human evaluation for the four dimensions. We ask evaluators to rate videos from each model over each dimension.

For semantic understanding dimensions (Spatial Understanding, Attribute Binding and Category Understanding), evaluation centers on instruction-following. The core rating rules is how accurately the subject in prompt is identified and animated to perform requested actions. Therefore, evaluators are asked to primarily focus on the consistency between the generated video and the prompt.
For reasoning dimension, rating rules focuses not only on whether the video accurately depicts the prompt, but more importantly, on its ability to perform reasonable casual inference based on the prompt and input image to generate outcomes consistent with real-world knowledge. To facilitate evaluation, we provide evaluators expected target outcomes for each reasoning case (the content is the same as "Prompt and Target" in Figure \ref{fig:case1},\ref{fig:case2}, \ref{fig:case3}) to assist scoring. Evaluators are asked to primarily focus on the consistency between the generated video and the provided expected outcomes.

For each generated video, evaluators assign scores on a 1–5 scale. Since human evaluation involves a high degree of subjectivity, we established scoring criteria for evaluators. (1) Very Poor: target subject is stationary. (2) Poor: target subject moves incorrectly, and other objects move as well. (3) Moderate: target subject moves incorrectly, but other objects remain still. (4) Good: target subject moves incorrectly, but other objects remain still. (5) Excellent: target subject moves correctly, and all other objects remain still.
% % % % % % % % % % % % % % % % % % % % % % % % % % % % % % % % % % 
% \clearpage
% \newpage
\section{Qualitative Comparison}
\begin{figure}[htbp]
    \centering
    \includegraphics[width=0.98\linewidth]{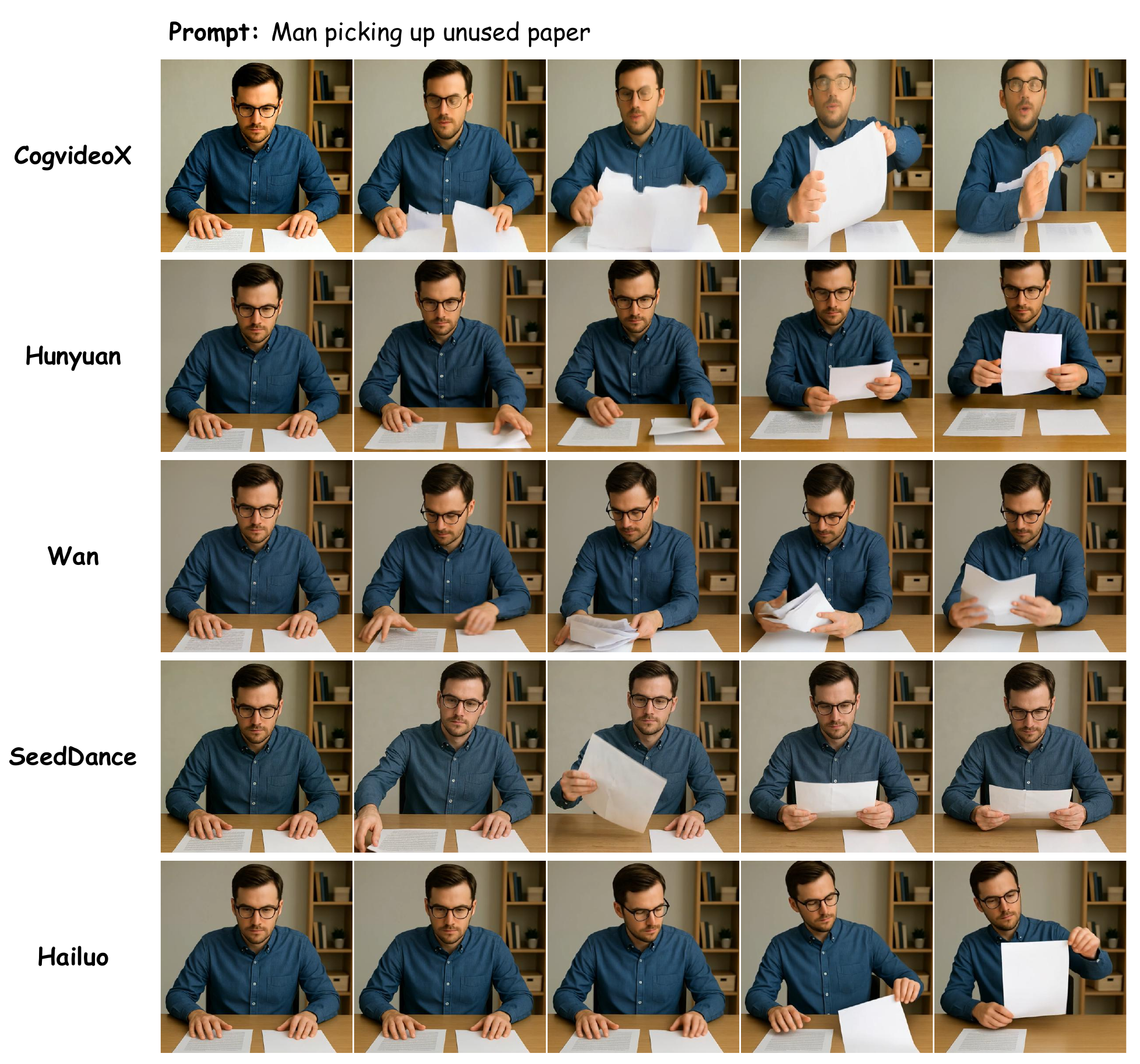} % 调整宽度为行宽的80%
    \caption{Qualitative Comparison on attribute.} % 图片标题
    \label{fig:qua1} % 图片标签，用于交叉引用
\end{figure} 
\clearpage
\newpage
\begin{figure}[htbp]
    \centering
    \includegraphics[width=0.98\linewidth]{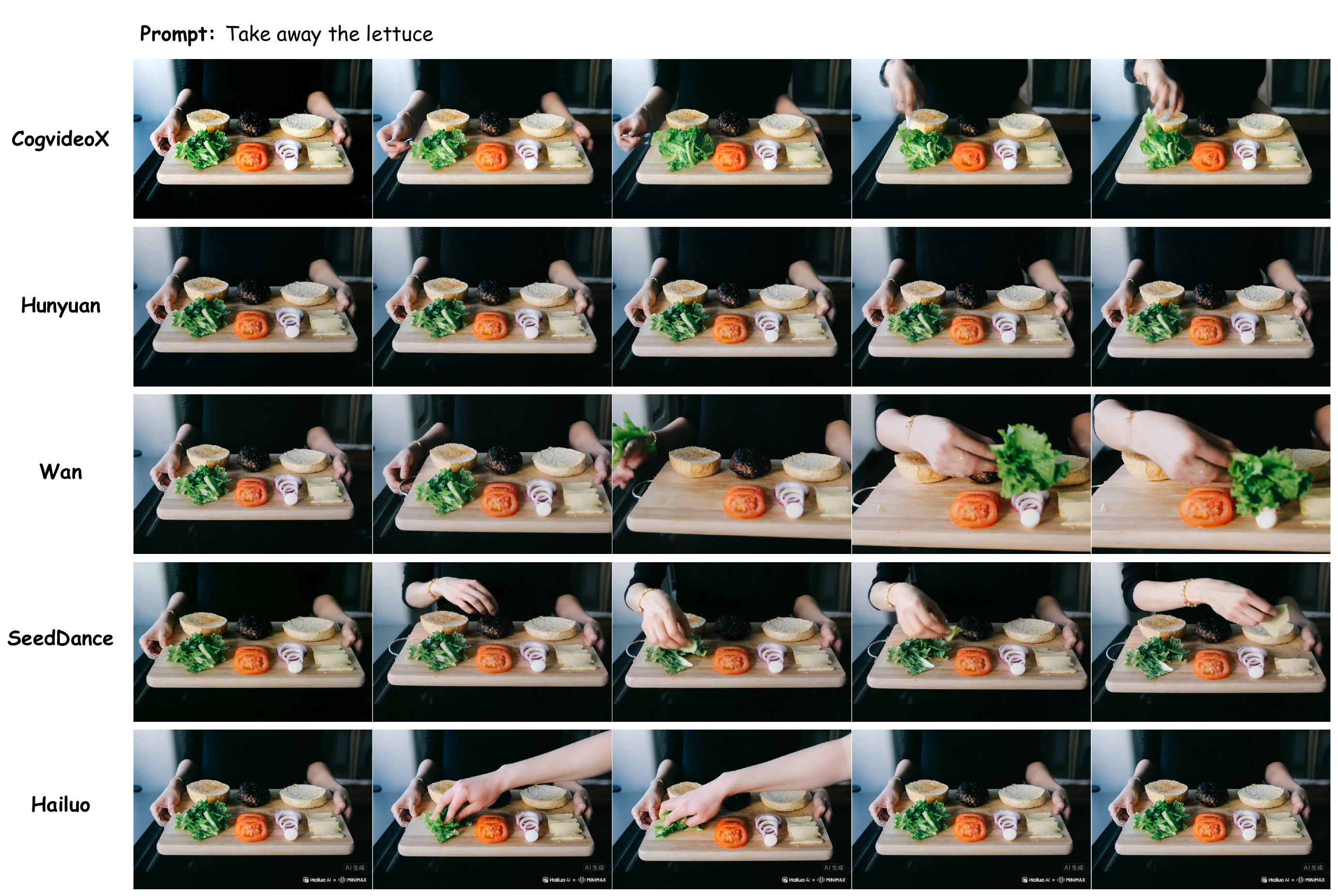} % 调整宽度为行宽的80%
    \caption{Qualitative Comparison on category.} % 图片标题
    \label{fig:qua1} % 图片标签，用于交叉引用
\end{figure} 

\clearpage
\newpage
\begin{figure}[htbp]
    \centering
    \includegraphics[width=0.98\linewidth]{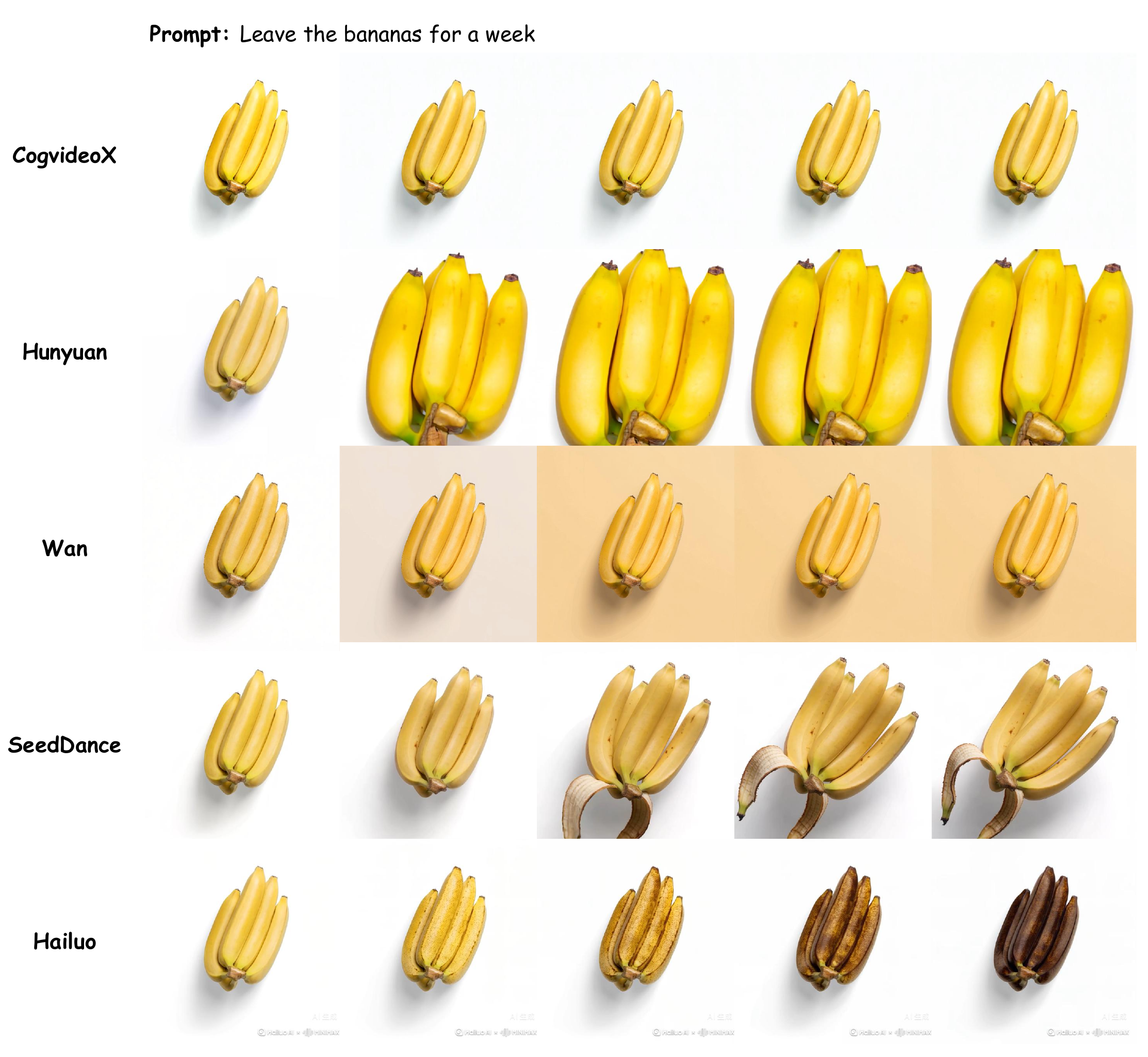} % 调整宽度为行宽的80%
    \caption{Qualitative Comparison on reasoning.} % 图片标题
    \label{fig:qua113} % 图片标签，用于交叉引用
\end{figure} 
\clearpage
\newpage
\begin{figure}[htbp]
    \centering
    \includegraphics[width=0.98\linewidth]{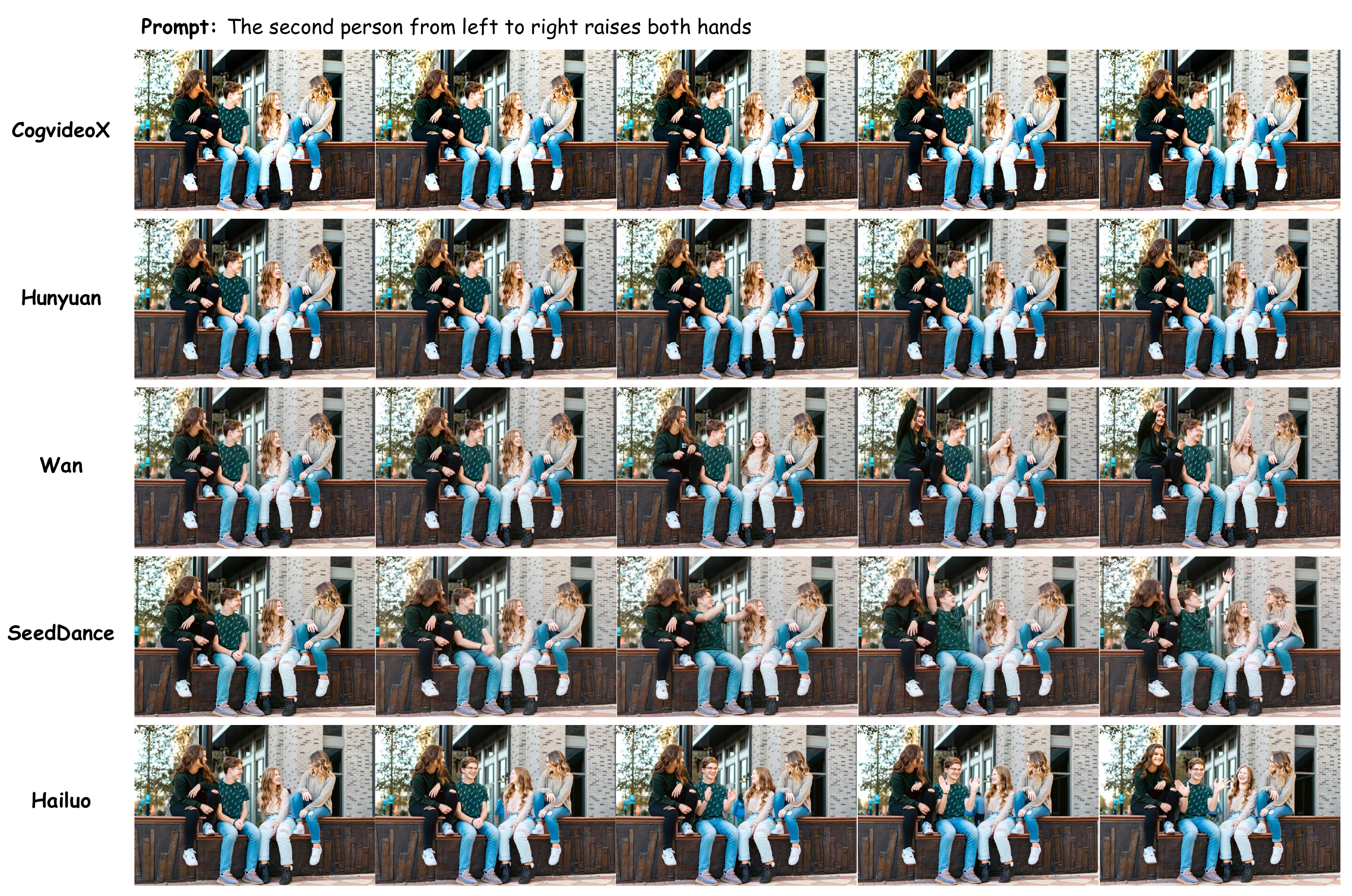} % 调整宽度为行宽的80%
    \caption{Qualitative Comparison on spatial.} % 图片标题
    \label{fig:qua1} % 图片标签，用于交叉引用
\end{figure} 

\clearpage
\newpage
\section{Limitation and Future Work}
While our work has taken a step forward by addressing a critical evaluation gap with our focus on semantics and reasoning, it still has certain limitations and remains room for improvement.
Firstly, the number of open-source I2V models is limited and their performance is relatively poor, while the more effective models are mostly commercialized and require payment. Due to the high cost for the API of commercial models, there are only two representative commercial close-sourced models are selected for evaluation on our benchmark. We will open-source our UI2VBench and encourage more I2V models to participate in the evaluation.
Secondly, we have only constructed cases for typical and common dimensions, and there is still room for expansion in this benchmark in the future. For example, for the spatial understanding dimension, we have only constructed linear arrangement. In the future, cases for subjects in complex arrangement such as circular sequence and grid array could be added. However, it is noteworthy that for almost all models requiring image understanding capabilities, comprehending complex spatial arrangements is a universal challenge. Even if cases involving complex spatial arrangements are constructed, accurately evaluating the generation outcomes based on existing methods would remain challenging.
Thirdly, the proposed evaluation metrics for sematic understanding place high demands on open-vocabulary object detectors, requiring them to accurately detect target subjects with the highest confidence. For example, accurately detecting "the second knife from the top" and "the sitting man" presents considerable challenges, which is crucial for precise detection.

\end{document}